\documentclass[10pt]{IEEEtran}
\IEEEoverridecommandlockouts
\usepackage{cite}
\usepackage{amsmath,amssymb,amsfonts}
\usepackage{algorithmic}
\usepackage{graphicx}
\usepackage{textcomp}
\usepackage{xcolor}
\usepackage[top=1in, bottom=0.75in, left=0.75in, right=0.75in]{geometry}
\usepackage[hidelinks]{hyperref}
\def\BibTeX{{\rm B\kern-.05em{\sc i\kern-.025em b}\kern-.08em
    T\kern-.1667em\lower.7ex\hbox{E}\kern-.125emX}}
\begin{document}

\title{\LARGE \textbf{Real-Time Neuromorphic Navigation: Integrating Event-Based Vision and Physics-Driven Planning on a Parrot Bebop2 Quadrotor}}

\author{Amogh Joshi, Sourav Sanyal  and Kaushik Roy \\
  \IEEEauthorblockA{Electrical and Computer Engineering,
 Purdue University\\
 \{joshi157, sanyals, kaushik\}@purdue.edu
 }
}
\maketitle

\begin{abstract}
In autonomous aerial navigation, real-time and energy-efficient obstacle avoidance remains a significant challenge, especially in dynamic and complex indoor environments. This work presents a novel integration of neuromorphic event cameras with physics-driven planning algorithms implemented on a Parrot Bebop2 quadrotor. Neuromorphic event cameras, characterized by their high dynamic range and low latency, offer significant advantages over traditional frame-based systems, particularly in poor lighting conditions or during high-speed maneuvers. We use a DVS camera with a shallow Spiking Neural Network (SNN) for event-based object detection of a moving ring in real-time in an indoor lab. Further, we enhance drone control with physics-guided empirical knowledge inside a neural network training mechanism, to predict energy-efficient flight paths to fly through the moving ring. This integration results in a real-time, low-latency navigation system capable of dynamically responding to environmental changes while minimizing energy consumption. We detail our hardware setup, control loop, and modifications necessary for real-world applications, including the challenges of sensor integration without burdening the flight capabilities. Experimental results demonstrate the effectiveness of our approach in achieving robust, collision-free, and energy-efficient flight paths, showcasing the potential of neuromorphic vision and physics-driven planning in enhancing autonomous navigation systems.

\noindent{Video Link: \url{https://youtu.be/9Gnjpb1k2Lo}}\\

\end{abstract}


\section{Introduction}
\vspace{-2mm}
Autonomous navigation algorithms have long been a focal point of interest in robotics research, with a recent emphasis on enhancing these systems using artificial intelligence (AI). However, existing AI-based navigation algorithms often struggle with reactive tasks like close-in obstacle avoidance, where rapid decisions are crucial. This limitation arises from the inability of traditional sensing modalities to provide necessary data quickly. Vision-based algorithms, while useful for high-level decision-making through techniques like object detection, are hindered by the high energy consumption and latency of conventional frame-based cameras, which can negate their obvious benefits.

\textit{Event-based cameras}\cite{dvs1,dvs2,dvs3} offer advantages over traditional \textit{frame-based cameras} in terms of latency and energy efficiency. They record only changes in intensity that exceed a programmable threshold. This capability significantly reduces bandwidth and allows for data capture at microsecond granularity. When these cameras are paired with \textit{spiking neurons} \cite{lif}, which process information asynchronously, they become highly effective for low-energy, latency-critical applications \cite{lee2020spike, kosta2023adaptive}. Together, they support the development of \textit{neuromorphic vision} and form the technological foundation of \textit{neuromorphic navigation algorithms}, facilitating energy-efficient and responsive autonomous navigation solutions \cite{sanyal2024ev}. Despite their potential, real-world implementations of these systems are quite rare, especially when it comes to their deployment on physical robots like drones and integrating them with physics-driven planning methods for complex operations.


In this work, we implement a real-time neuromorphic navigation algorithm on a Parrot Bebop2 quadrotor, utilizing event-based neuromorphic vision and physics-driven planning to autonomously navigate through a moving ring in an indoor environment while avoiding collisions.\footnote{Code available here: \url{https://github.com/amoghj98/neuroNav}}

\begin{figure}[!t]
\begin{center}
   \includegraphics[width = 0.37\textwidth]
   {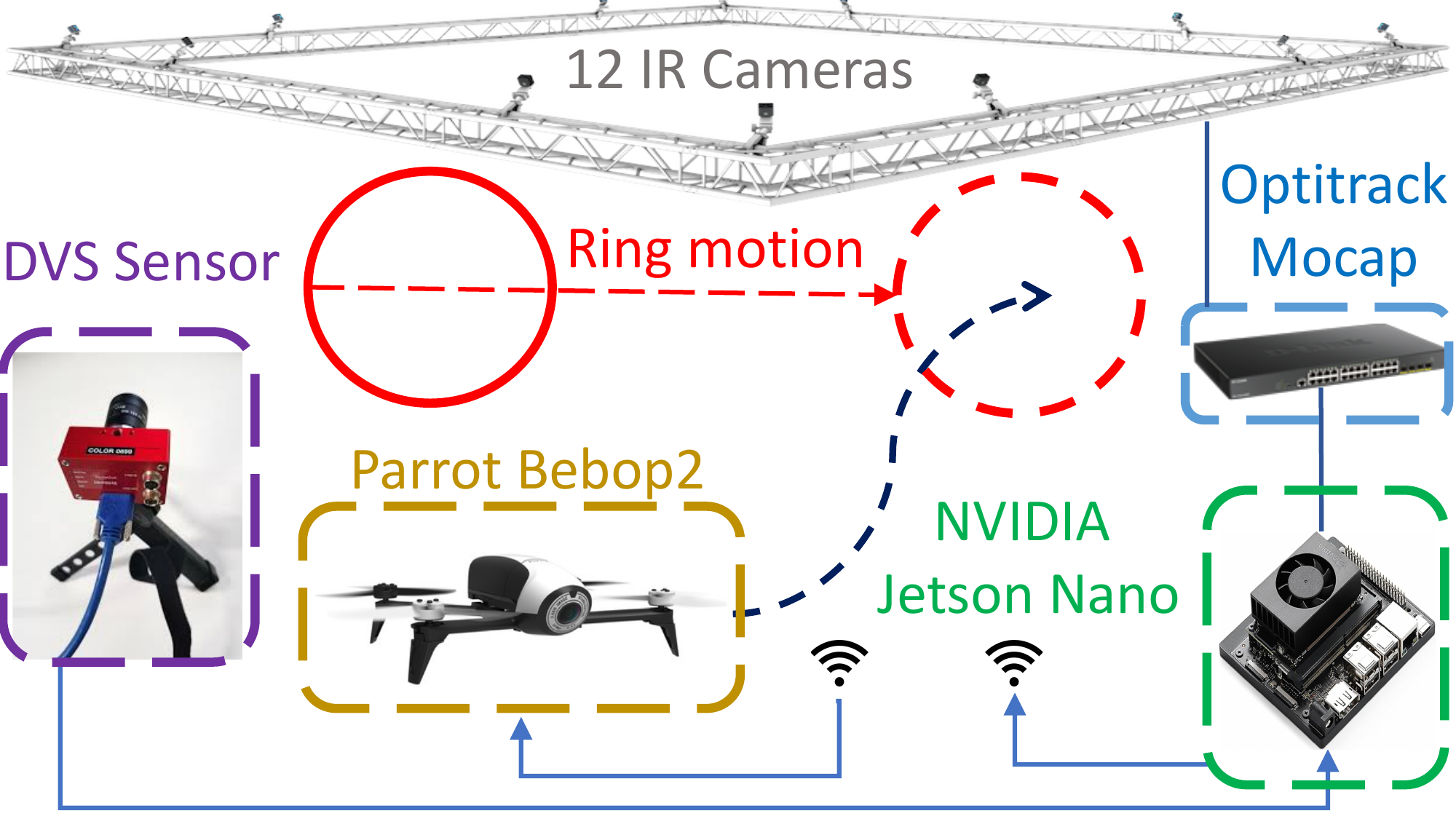} 
   \label{method}
   \caption{High-level overview of our real-time neuromorphic navigation setup. Drone and DVS sensor poses are taken from the Optitrack motion capture system consisting of 12 IR cameras. The ring is tracked using the DVS Sensor. The algorithm is executed on an Off-Board NVIDIA Jetson Nano edge processor to publish control commands to the Parrot Bebop2 over a private WiFi network.}
\end{center}
\vspace{-4mm}
\end{figure}
\section{Background and Related Work}

A key aspect of neuromorphic vision is its use of \textit{spiking neurons} arranged spatially to form a Spiking Neural Network (SNN) for object detection. The DOTIE framework by Nagaraj et al. \cite{nagaraj2023dotie}, which isolates events based on object speed, has shown promise in improving the speed and accuracy of object detection tasks.

In recent years, physics-based AI has proven powerful in embedding physical system knowledge into neural network learning processes, enhancing model robustness and adaptability. RAMP-Net \cite{sanyal2023ramp} uses physics-informed neural networks to boost control accuracy and adaptability in quadrotors. Similarly, KNODE-MPC by Chee et al. \cite{chee2022knode} employs a knowledge-based control framework that integrates empirical data with physical laws to enhance aerial robot performance.

Building on these innovations, the EV-Planner framework \cite{sanyal2024ev} merges neuromorphic vision sensors and physics-guided neural networks, addressing traditional sensing delays and optimizing actuator energy use. This integration enhances endurance and efficiency in resource-constrained environments.
\section{System Overview and Hardware Implementation}
In this work, we implement a real-time version of the neuromorphic navigation algorithm EV-Planner \cite{sanyal2024ev} on a quadrotor drone using event-based vision and physics-driven planning. We use low-level Proportional-Integral-Derivative (PID) controllers to execute the plan output by EV-Planner in the real world. We call this real-world implementation \textit{EV-PID}. Using our engineered setup, the drone autonomously navigates through a moving ring in an indoor environment, while avoiding collision. To highlight the robustness of our implementation, the motion of the ring is controlled using a teleoperated TurtleBot moving at a random velocity.

Considering the constraints of our indoor space and to enhance safety, we selected the Parrot Bebop2 quadrotor as our platform. The Bebop2's Max. Takeoff Weight (MTOW) of $\sim600$gm is insufficient to lift the combined weight of the drone, the DVS sensor, and the Jetson nano compute board. Consequently, we mounted the DVS sensor on a tripod and used an external Optitrack motion capture system \cite{optitrack} to account for movement between the drone and the sensor. Larger drones, which could carry both the DVS sensor and Jetson, do not require such compensation but are unsuitable for safe flight in our confined indoor space.

Fig. \ref{method} provides an overview of our real-time neuromorphic navigation setup. 
The ring target is mounted on an overhead cable and moved by a teleoperated TurtleBot using pulleys rigged in a double-tackle configuration (see the video), enabling bi-directional motion at speeds $5$cm/s - $50$cm/s.


Initially, the drone and the teleoperated TurtleBot are armed; the drone then takes off and stabilizes. Once stable, the TurtleBot begins to move, prompting the drone to start tracking the ring. To test the robustness of our system, we randomize the ring's velocity and the drone's maneuver start time. Positional data from the drone and the DVS sensor, captured using an external motion-capture system, are combined with DVS readings and relayed to the Jetson via Ethernet for motion compensation. The EV-Planner algorithm on the Jetson processes this data to generate setpoints, which are then converted to roll, pitch, yaw, and thrust commands. These commands are transmitted over a private WiFi network and executed by the drone's low-level PID controller.

\section{Results and Discussion}
\begin{figure}[!t]
\begin{center}
   \includegraphics[width = 0.34\textwidth]
   {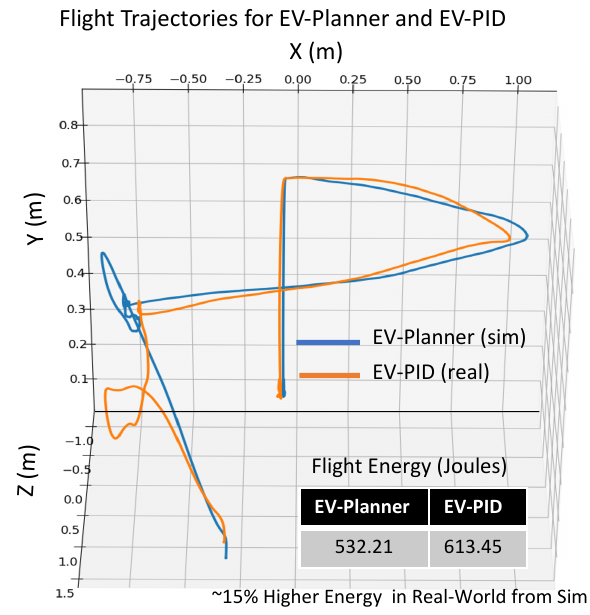} 
   \caption{Flight Trajectories of EV-Planner (simulation) and EV-PID (real-world implementation) while flying through the moving ring without collision. Deviations from the EV-Planner's planned path due to controller artifacts and motor non-idealities are clearly visible which is also reflected in the flight energy values with $\sim15\%$ higher actuation energy observed for EV-PID compared to EV-Planner.}
   \label{test}
\end{center}

\end{figure}

Figure \ref{test} illustrates the trajectories for both simulated and real-time runs of the neuromorphic navigation algorithm. The initial randomness in both flight trajectories reflects the additional measures we took to ensure robustness. These include performing flat-trim and alt-hold and maintaining positional and velocity bounds to ensure safety. The flight path traced by EV-PID (real-world implementation) deviates from the path traced by EV-Planner (simulation) primarily due to the inherent non-idealities in motor control and physical constraints not present in the simulation. We also observed the flight and motion paths for the EV-PID framework to be sharper and more abrupt as compared to EV-Planner. We present the actuation energies of the two methods using the energy model of the brushless DC motor reported in \cite{sanyal2024ev}.
The energy consumption of the real-world EV-PID flight is found to be $\sim15\%$ higher than its simulated counterpart (EV-Planner). This increase in energy highlights the challenges and inefficiencies that emerge when transitioning from simulation to actual physical environments. Please note that the energies reported are actuation energies and not the significantly lower perception compute energies.
\section{Conclusion}

The study reveals a slight disparity between simulated predictions and the real-time performance of the neuromorphic navigation system. The effectiveness of combining neuromorphic vision with physics-based planning was demonstrated for collision avoidance in flight. The EV-PID system's higher energy usage in real-world applications underscored the impact of environmental factors and mechanical limitations, which are often overlooked in prototypical simulations. Future research to optimize these systems further can potentially explore advanced sensor integration and algorithmic refinements, thus minimizing the energy discrepancies observed between simulation and the real world.

\bibliographystyle{IEEEtran}
\bibliography{conference_101719}


\end{document}